\pdfoutput=1

\documentclass[11pt]{article}

\usepackage{emnlp2021}

\usepackage{times}
\usepackage{latexsym}

\usepackage[T1]{fontenc}
\usepackage{times}
\usepackage{latexsym}
\usepackage{graphicx}
\usepackage{amsmath}
\usepackage{amssymb}
\usepackage{amsfonts}
\usepackage{bbm}
\usepackage{multirow}
\usepackage{tabularx}
\usepackage{ragged2e}
\usepackage{booktabs}
\usepackage{adjustbox}
\usepackage{diagbox}
\usepackage{pifont}
\usepackage{makecell}
\usepackage{subfigure}
\usepackage{microtype}
\usepackage{todonotes}
\usepackage{kotex}

\usepackage{color}
\usepackage{dirtytalk}
\usepackage{multirow}
\def\ie{\textit{i.e.}}
\def\eg{\textit{e.g.}}

\newcommand\ttt[1]{\texttt{#1}}

\newcommand{\cls}{[\textsc{cls}]}
\newcommand{\sep}{[\textsc{sep}]}

\newcommand{\del}{[\textsc{del}]}

\newcommand{\ui}{[\textsc{ui}]}
\newcommand{\mask}{[\textsc{mask}]}
\DeclareMathOperator*{\argmax}{argmax}
\newcommand{\ur}{[\textsc{ur}]}

\usepackage[utf8]{inputenc}

\usepackage{microtype}

\title{Self-supervised Contrastive Cross-Modality Representation Learning for Spoken Question Answering}

\author{Chenyu You$^{\dagger*}$ \quad Nuo Chen$^{\ddagger*}$ \quad Yuexian Zou$^{\ddagger\mathsection}$ \\
$^{\dagger}$ Department of Electrical Engineering, Yale University, New Haven, CT, USA\\
$^{\ddagger}$ADSPLAB, School of ECE, Peking University, Shenzhen, China\\
$^{\mathsection}$Peng Cheng Laboratory, Shenzhen, China\\
\ttt{chenyu.you@yale.edu}\\
\ttt{\{nuochen,zouyx\}@pku.edu.cn}
}

\begin{document}
\maketitle
\renewcommand{\thefootnote}{\fnsymbol{footnote}}
\footnotetext[1]{Equal contribution.}
\renewcommand{\thefootnote}{\arabic{footnote}}

\begin{abstract}
Spoken question answering (SQA) requires fine-grained understanding of both spoken documents and  questions for the optimal answer prediction. In this paper, we propose novel training schemes  for spoken question answering with a self-supervised training stage and a contrastive representation learning stage. In the self-supervised stage, we propose three auxiliary self-supervised tasks, including \textit{utterance restoration}, \textit{utterance insertion}, and \textit{question discrimination}, and jointly train the model to capture consistency and coherence among speech documents without any additional data or annotations. We then propose to learn noise-invariant utterance representations in a contrastive objective by adopting multiple augmentation strategies, including \textit{span deletion} and \textit{span substitution}. Besides, we design a Temporal-Alignment attention to semantically align the speech-text clues in the learned common space and benefit the SQA tasks. By this means, the training schemes can more effectively guide the generation model to predict more proper answers. Experimental results show that our model achieves state-of-the-art results on three SQA benchmarks.

\end{abstract}

\begin{figure*}[t]
    \centering
    \includegraphics[width=1.0\linewidth]{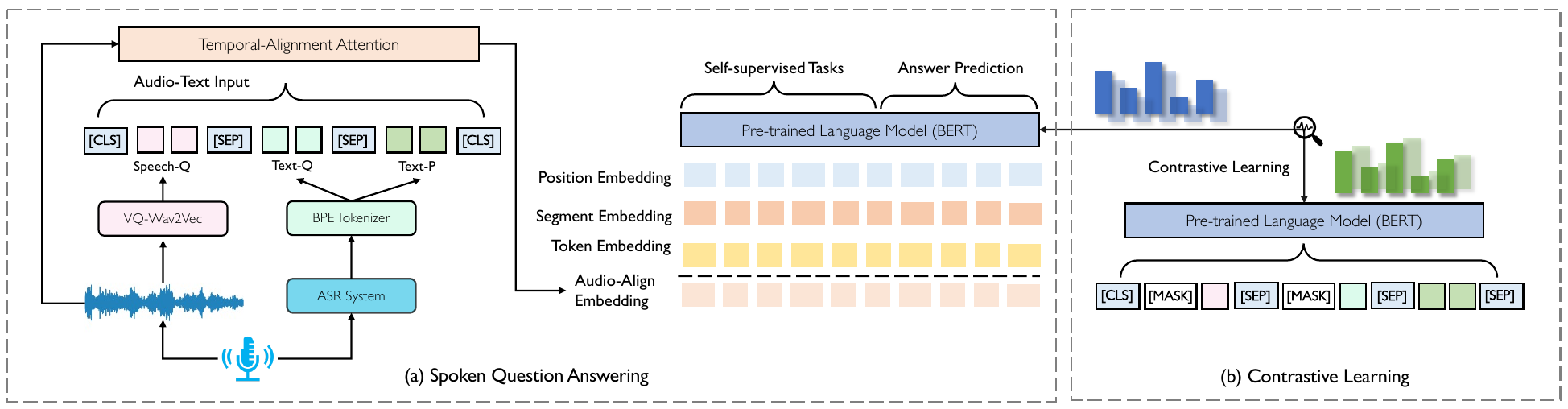}
    \vspace{-15pt}
    \caption{Overall architecture of our model: (a) For a spoken QA part, we use VQ-Wav2Vec and Tokenizer to transfer speech signals and text to discrete tokens. A Temporal-Alignment Attention mechanism is introduced to match each text embedding with the corresponding speech features. Then, we use BERT to learn sequential information of utterances with the proposed self-supervised tasks. We generate the final answer distribution on both domains. At inference time, we use the BERT only. (b) We incorporate contrastive learning strategies to train our SQA model in an auxiliary manner to improve the model performance.}
    \vspace{-10pt}
    \label{fig:my_label}
\end{figure*}

\section{Introduction}

Building an intelligent spoken question answering (SQA) system has attracted considerable attention from both academia and industry. In recent years, many significant improvements have achieved in speech processing and natural language processing (NLP) communities, such as multi-modal speech emotion recognition \cite{beard2018multi,sahu2019multi,priyasad2020attention,siriwardhana2020jointly}, spoken language understanding \cite{mesnil2014using,chen2016end,chen2018spoken,haghani2018audio}, and spoken question answering \cite{li2018spoken,you2020data,you2020contextualized,you2021knowledge}. Among these topics, SQA is an especially challenging task, as it requires the machines to fully understand the semantic meaning in both speech and text data, and then provide the correct answer given a question and corresponding speech documents. 

Automatic speech recognition (ASR) and text question answering (TQA) are two key components to build such a SQA system. The former module is used for transforming the speech sequences into text form, and the latter module trained on noisy ASR transcriptions utilizes NLP techniques to give a concrete answer. However, utilizing existing state-of-the-art SQA systems to retrieval answers still remain formidable challenges, such as ASR recognition errors. This is mainly because ASR systems usually fail to recognize the speech, leading to word errors (\eg, \say{Barcelona} to \say{bars alone}).

To address these issues, most existing SQA methods are either text-based \cite{li2018spoken,lee2018odsqa,lee2019mitigating,chuang2019speechbert} or fusion-based \cite{you2020contextualized,you2020data,you2021knowledge}. One line of research examines internal vector representations both in speech and text domains \cite{li2018spoken,lee2019mitigating}, often using sub-word units for language modeling. Another line of work \cite{you2020contextualized,you2021knowledge} investigates the transfer learning problem about how to leverage a large amount of speech and text data to improve the performance of SQA. However, some critical challenges remain, such as robustness, generalization, and data efficiency.

Different from previous methods \cite{su2020improving,li2018spoken,lee2019mitigating,you2021knowledge}, we move beyond leveraging dual nature of TQA and ASR to mitigate recognition errors. In this paper, we focus not only on extracting the cross-modality information for joint spoken and textual understanding, but also on the training procedure that may take the most advantage of the given dataset. Inspired by the recent advance in contrastive learning \cite{chen2020simple,khosla2020supervised} and recent breakthrough \cite{devlin2018bert,liu2019roberta,rahman2020integrating,chen2020adaptive} in the context of NLP, we propose a novel training framework for Spoken QA that integrates these two perspectives to improve spoken question answering performance. Our training framework contains two steps: (a) self-supervised training stage, and (b) contrastive training stage. During the self-supervised training stage, instead of building the complex spoken question answering model, we propose to learn a spoken question answering system based on pre-trained language models (PLMs) with several auxiliary self-supervised tasks. 
In particular, we introduce three self-supervised tasks, including \textit{utterance restoration}, \textit{utterance insertion}, and \textit{question discrimination}, and jointly train the model with these auxiliary tasks in a multi-task setting. On the one hand, these auxiliary tasks enable the model to capture sequential order within the given passage. On the other hand, they effectively learn cross-modality knowledge without any additional dataset or annotations to generate better representations for answer prediction. 

During the fine-tuning stage, along with the main QA loss, we incorporate the contrastive learning strategy to our framework in an auxiliary manner for the SQA tasks. Specifically, we use multiple  augmentation strategies, including \textit{span deletion} and  \textit{span substitution}, to develop the capability of learning noise-invariant utterance representations. In addition, we propose a novel attention mechanism, termed Temporal-Alignment Attention, to effectively learn cross-modal alignment between speech and text embedding spaces. By this mean, our proposed attention mechanism can encourage the training process to pay more attention to semantic relevance, consistency and coherency between speech and text in their contexts to provide better cross-modality representations for answer prediction. The overview of our framework is shown in Figure \ref{fig:my_label}. We evaluate the proposed approach on the widely-used spoken question answering benchmark datasets - Spoken-SQuAD \cite{li2018spoken}, Spoken-CoQA \cite{you2020data}, and 2018  Formosa Grand Challenge (FGC). Experimental results show our proposed approach outperforms other state-of-the-art models when self-supervised training is preceded. Moreover, evaluation results indicate our learning schema can also consistently bring further improvements to the performance of existing methods with contrastive learning.

\begin{figure*}[t]
    \centering
    \includegraphics[width=1.0\linewidth]{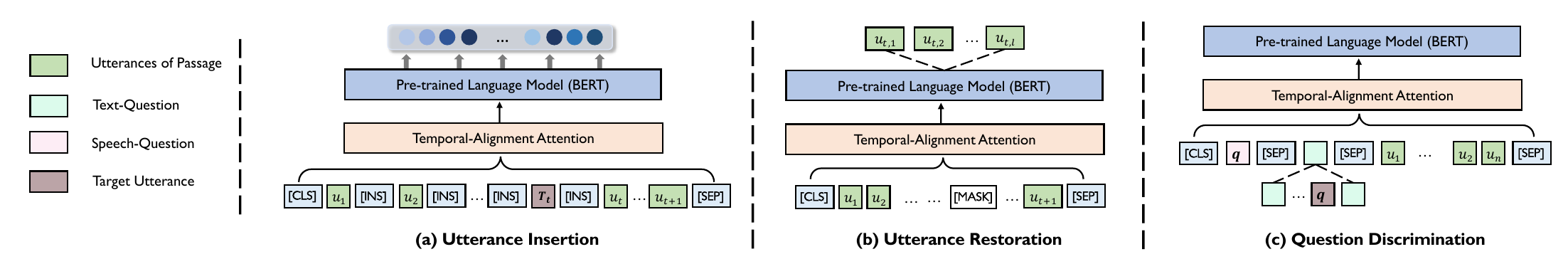}
    \vspace{-15pt}
    \caption{Auxiliary tasks in Self-supervised training.}
    \label{fig:self}
    \vspace{-10pt}
\end{figure*}

\section{Related Work}
\paragraph{Spoken Question Answering.}

Spoken question answering \cite{li2018spoken,lee2018odsqa,lee2019mitigating,su2020improving,huang2021audio,you2020contextualized,you2021knowledge,you2020data,you2021mrd,chen2021self} is a task of generating meaningful and concrete answers in response to a series of questions from spoken documents. Typical spoken QA systems focus on integrating ASR and TQA in one pipeline. ASRs are designed to transcribe audio recordings into written transcripts. However, current ASRs are not capable of processing every spoken document. Generated ASR transcripts may contain highly noisy data, which severely influences the performance of QA systems on speech documents. A number of works have explored the shortcomings of this issue. \citet{li2018spoken} and \citet{lee2018odsqa} introduced sub-word unit strategy to alleviate the effects of speech recognition errors in SQA. SpeechBERT \cite{chuang2019speechbert} utilized the pre-trained BERT-based language model to effectively learn audio-text features. The model improved the performance of ASR by SpeechBERT. However, these works mainly focus on improving performance by exploiting internal information without considering learning the explicit mapping between human-made transcripts and corresponding ASR transcriptions, which is crucial to building Spoken QA systems. \citet{lee2019mitigating} adopted an adversarial learning strategy to alleviate this gap to achieve remarkable performance improvements. In contrast to previous works in SQA, which only consider speech representations or confine to certain subtasks (\eg, spoken multi-choice question answering and spoken conversational question answering), we not only model the interactions between speech and text data, but also focus on capturing semantic similarity. In parallel, our proposed method is a unified framework, which can be easily applied to a variety of downstream speech processing tasks.

\paragraph{Self-supervised Learning.}
Self-supervised learning (SSL) has become a promising solution for performance improvements by leveraging large amounts of unlabeled audio data. Substantial efforts have recently been dedicated to developing powerful SSL-based approaches in the machine learning community. \cite{oord2018representation,you2018structurally,schneider2019wav2vec,baevski2019vq,you2019low,you2019ct,chung2019unsupervised,pascual2019learning,liu2020mockingjay,chung2021splat,you2020unsupervised,you2021momentum,you2021simcvd}. \citet{oord2018representation} designed a Contrastive Predictive Coding (CPC) framework to learn compact latent representations to provide future predictions over future observations by combining autoregressive modeling and noise-contrastive estimation in an unsupervised manner. Later on, \citet{schneider2019wav2vec} further applied the learned generic speech representations to improve supervised ASR systems. \citet{chung2019unsupervised} and \citet{liu2020mockingjay} have taken advantage of state-of-the-art self-supervised pre-trained language models in the NLP community. These methods mainly focus on learning from audio data only, yet hardly exploit meaningful and relevant representations across both speech and text domains. Most recently, \citet{khurana2020cstnet} investigated how to leverage speech-translation retrieval tasks into self-supervised learning. In this study, we explore an effective way to utilize cross-modality information via the self-supervised training scheme for SQA tasks without additional large-scale unlabeled datasets. In contrast, our proposed method yields such remarkable accuracy without using any extra data or annotations.

\paragraph{Contrastive Representation Learning.}

In parallel to self-supervised learning, an emerging subfield has explored the prospect of contrastive representation learning in the machine learning community \cite{kharitonov2021data,manocha2021cdpam,oord2018representation,he2020momentum,chen2020simple,hjelm2018learning,tian2019contrastive,henaff2020data,wu2018unsupervised,khurana2020cstnet}. This is often best understood as follows: pull together the \textit{positive} and an anchor in embedding space, and push apart the anchor from many \textit{negatives}. Thus, the choice of \textit{negatives} can significantly determine the quality of the learned latent representations. Since contrastive learning is a framework to learn representations by comparing the similarity between different views of the data. In computer vision, \citet{chen2020improved} has demonstrated that the enlarged negative pool significantly enhances unsupervised representation learning. However, there are few attempts on contrastive learning to address downstream language processing tasks. Recently, few prior work \cite{kharitonov2021data} incorporated CPC with time-domain data augmentation strategies into contrastive learning framework for speech recognition tasks. In contrast, we focus on learning interactions between speech and text modalities for spoken question answering tasks, and also introduce a set of auxiliary tasks on top of the former self-supervised training scheme to improve representation learning.

\section{Methods}

In this section, we first formalize the spoken question answering tasks. Furthermore, we introduce the key components of our method with self-supervised contrastive representation learning. Next, we describe the design of our proposed Temporal-Alignment Attention mechanism. Lastly, we discuss how to incorporate contrastive loss into our self-supervised training schema.

\subsection{Task Formulation}

Suppose that there is a dataset $\mathcal{D} \in\{Q_i,P_i,A_i\}_{i}^{N}$, where $Q_i$ denotes a question, $P_i$ denotes a passage with a answer $A_i$. In this study, similar to the SQA setting in \cite{lee2018odsqa,kuo2020audio}, we focus on extraction-based SQA, which can be applied to other types of language tasks. We use Spoken-SQuAD, Spoken-CoQA, and FGC to validate the robustness and generalization of our proposed approach. In Spoken-SQuAD, $Q_i$ and $A_i$ are both single sentences in text form, and $P_i$ consists of multiple sentences in spoken form. In FGC, $Q_i$, $A_i$, and $P_i$ are all in spoken form. Different from Spoken-SQuAD and FGC, Spoken-CoQA is in a multi-turn conversational SQA setting, which is more challenging than a single-turn setting. Moreover, it adopts $Q_i$ in spoken form. The task is to learn a SQA model $\mathrm{G}(\cdot,\cdot)$ from $\mathcal{D}$ so that $\mathrm{G}(Q_{i},P_{i})$ can provide a most proper answer $A_i$ to the given question $Q_i$.

\subsection{Spoken question answering with PLMs.}
\label{sec:PLMs}

Recent PLMs, such as BERT \cite{devlin2018bert} and ALBERT \cite{lan2019albert}, learn meaningful language representations from large amounts of unstructured corpora, and have achieved superior performances on a wide range of downstream tasks in the domain of NLP. Following previous work \cite{lee2019mitigating}, we consider building the SQA system with PLMs. We adopt BERT as the base model for a fair comparison. Similar to \citet{lee2018odsqa}, we concat ASR token sequences of a passage and a question as input to our SQA system. Specifically, given a passage $P\!\!=\!\!\{p_1,p_2,...,p_n\}$ and a question $Q=$
$\{q_1,q_2,...,q_m\}$, we first concatenate all utterance sequences, which can be formulated as $\mathbf{X}\!\!=\!\!\{\cls,q_1,q_2,...,q_m,\sep,p_1,p_2,...,p_n,\sep\}$. \say{$\cls$} and \say{$\sep$} denote begin token and separator token of each concatenated token sequence, respectively. We then utilize the pre-trained BERT to extract the hidden state features from the processed token sequences. Finally, we feed these 
representations to the following module, including a feed-forward network followed by a softmax layer, to obtain the probability distribution for each answer candidate given a textual passage-question pair. We use the cross-entropy loss as the question answering loss.

\subsection{Self-supervised Training}
\label{subsec:selfsupervised}
Heading for a SQA model that can effectively make use of cross-modality knowledge with a limited number of training data and produce better contextual representations for answer prediction. To this end, we design three auxiliary self-supervised tasks, including \textit{utterance restoration}, \textit{utterance insertion}, and \textit{question discrimination}. The objective of these auxiliary tasks is to capture the semantic relevance, coherence, and consistency between speech and text domains. Figure \ref{fig:self} illustrates three auxiliary self-supervised tasks. These tasks are jointly trained with the SQA model in a multi-task manner. More training examples of self-supervised training can be found in Table \ref{tab:masked_input} and Appendix Table \ref{table:task1ex-canada}.

\paragraph{Utterance Insertion.}

PLMs often suffer from the limitations in capturing latent semantic and logical relationships in discourse-level, which refers to the problem that Next Sentence Prediction (NSP), the standard training objective of PLM-based approaches, negatively impact semantic topic shift without modeling coherence. One key reason is that NSP fails to capture sufficient semantic coherence with a incomprehensible passage \cite{lan2019albert}, which leads performance degradation. Thus, learning the natural sequential relationship between consecutive utterances within a passage can significantly help the model understand the meaning of the passage. 

In order to solve the above-mentioned problem, we design a more general self-supervised task with the spoken question answering context termed \textit{utterance insertion}. In this way, it can enable the model to fully leverage the sequential relationship within a passage to improve the performance in calculating the semantic relevance between consecutive utterances. specifically, we first extract $k$ consecutive utterances from one passage. Then we insert an utterance, which is randomly selected from another topic unrelated passage. Hence, suppose $k+1$ utterances consist of $k$ utterances from the original passage and one from different corpus, the goal is to predict the position of inserted utterance given the $k+1$ utterances. A special token $\ui$ is introduced to be positioned before each utterance. The input can be formulated as follows:
\begin{equation}
 \begin{split}
\mathbf{X}_{\textsc{ui}} = [\cls\,\ui_1\,u_1\ui_2\,...\,\ui_t\,u_{\textsc{ins}}\qquad\\
\ui_{t+1}\,u_{t}\,...\,\ui_{k+1}\,u_{k}\,\sep],
\end{split}
\end{equation}
where $u_{\textsc{ins}}$ is the inserted utterance.

\paragraph{Utterance Restoration.}
One of the major tasks to train PLMs is mask token prediction (MTP), which requires the model to estimate the position of the masked utterance during the training stage. Although, recent work \cite{liu2019roberta,lan2019albert,devlin2018bert,joshi2020spanbert} found that utilizing this auxiliary task can improve model performance, it only focuses on learning syntactic and semantic representations of the word in token-level. However, spoken question answering is a more challenging task, which requires the deeper understanding of each utterance within a passage. To explicitly model the utterance-level interaction between utterances within a passage, we propose an utterance-level masking strategy termed \textit{utterance restoration} to predict the utterance, which causes inconsistency. Specifically, suppose that a context is $c=\{u_1,u_2,...,u_k\}$ including $k$ consecutive utterances, we first randomly pick an utterance $u_t, t \!\in\! [0,k]$, and then replace all tokens in the $u_t$ by using a special token $\mask$. Similarly, a special token $\ur$ is introduced to be positioned before each utterance. To adapt the task in BERT, we formulate input of BERT encoder as follows:
\begin{equation}
 \begin{split}
\mathbf{X}_{\textsc{ur}} = [\cls\,\ur_1\ u_1...\,\ur_t\ u_{\textsc{mask}}\, \qquad\\
\ur_{t+1}\ u_{t+1}...\,\ur_{k}\ u_{k}\sep],
\end{split}
\end{equation}
where $u_{\textsc{mask}}$ consists of only $\mask$ tokens, which has the same length with $u_t$.

\begin{figure}[t]
    \centering
    \includegraphics[width=0.95\linewidth]{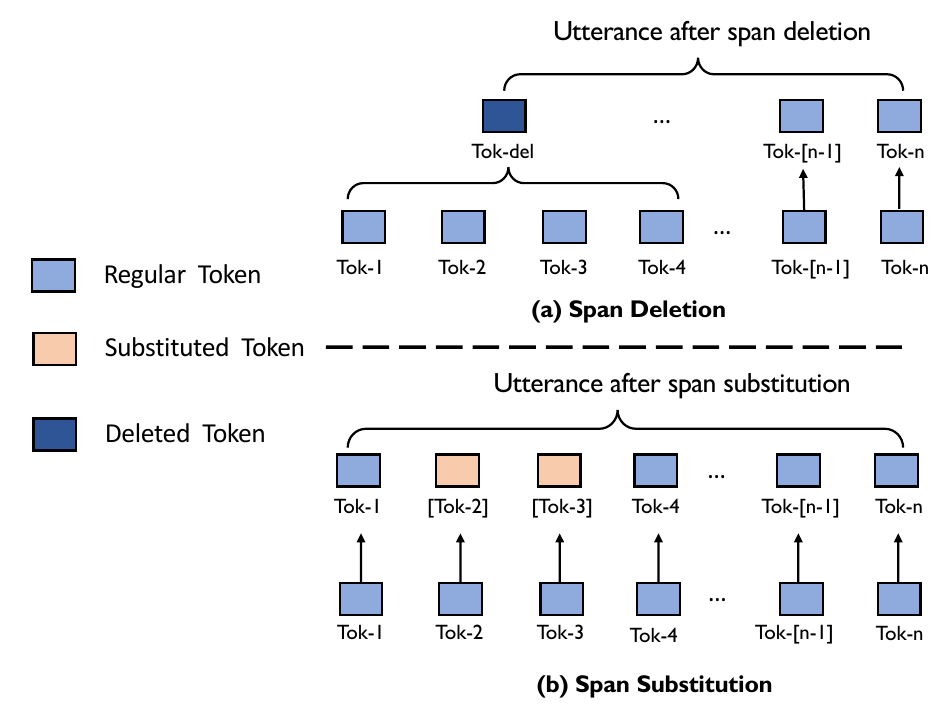}
    \vspace{-10pt}
    \caption{Auxiliary tasks in Contrastive Learning.}
    \label{fig:contras}
    \vspace{-10pt}
\end{figure}

\paragraph{Audio-Text Input.}
Inspired by recent success in video question answering \cite{kim2020self}, we leverage the cross-modality sequence modeling to generate audio-text sequence as input for \textit{question discrimination} task. In this process, we utilize the BPE tokenizer to convert the ASR documents to a sequence of \textit{Text-Question} and \textit{Text-Passage} tokens, similar to PLMs (See Section \ref{sec:PLMs}). We utilize pre-trained VQ-Wav2Vec \cite{baevski2019vq} trained on Librispeech-960 \cite{panayotov2015librispeech} to encode speech signals to a sequence of input tokens for \textit{Speech-Question}, since it outperforms the conventional RNN/CNN on sequence modeling.

\paragraph{Question Discrimination.}
Recent work \cite{kuo2020audio} has shown that learning cross-modality representation is essential for SQA tasks. Hence we design \textit{question discrimination} to consider building semantic alignments between speech and text by incorporating cross-modality knowledge into our model. Unlike the original goal of SQA (\ie, finding the answer using a question and contextualized contexts in Section \ref{sec:PLMs}), we instead train the model to predict the proper text question using audio-text contexts. Specifically, we first randomly select $k-1$ questions in textual form from other passages, and then incorporate them into the corresponding question $Q_t$. We can reformulate the question as $\hat{Q}=\{Q_t^{1},..,Q_t^{k-1},Q_t\}$. The goal of this task is to find the correct \textit{Text-Question} given a \textit{Speech-Question} and \textit{Text-Passage} contexts.
\vspace{-1mm}
\begin{equation}
    \bar{Q}=\argmax \mathbb{P} (Q_i|Q_s,P), Q_i\in \hat{Q},
    \vspace{-5pt}
\end{equation}
where $Q_s$ denotes the appropriate question in spoken form.

\subsection{Temporal-Alignment Attention}
Our proposed Temporal-Alignment Attention strategy is in the spirit of selectively leveraging cross-modality knowledge for SQA. Given an ASR token $\mathcal{U}^{i}$ and its corresponding acoustic-level MFCC features $F^{i}$, the goal is to enhance the SQA model by learning semantically meaningful alignment between speech and text domain \footnote{$\mathcal{U}^{i}$ can be any token in $P_i$ and $Q_i$. Similar to \cite{kuo2020audio}, for each acoustic frame, we use 40 MFCCs obtained from 40 FBANKs with 3 pitch features as input for ASR module and for our model.}. To align speech and text embeddings, we use a simple fully-connected feed-forward layer. The speech embedding features $\hat{r}^{i}$ is processed by self-attention to obtain speech-aligned features. Formally, the proposed attention module is defined as follows:
\vspace{-1mm}
\begin{equation}\label{TAtt}
\begin{split}
&r^{i}=\sum_{i=1}^{|\mathcal{U}_i|}[\,\text{softmax}(W^{i}F^{i}) * F^{i}\,]_j,\\
&\hat{r}^{i}=\text{FNN}(r^{i}),\\
&\{\textbf{u}^i\}=\text{Attention}(\hat{r}^{i},\hat{r}^{i},\hat{r}^{i}),
\end{split}
\vspace{-5mm}
\end{equation}
where $W^{i}$ is parameters. $*$ denotes element-wise multiplication. $[\cdot]_j$ is $j$-th column of a matrix. $\hat{r}^{i}$ and $\text{Attention}$ are acoustic-level embedding and self-attention, respectively. Note that we set $\textbf{u}^i$ of each special token (\eg, \cls) to 0.

\begin{table}[t]
\resizebox{\linewidth}{!}
{
\small
\begin{tabular}{p{3.6in}}
	\toprule
		\textbf{Original text Input} \\
	\vspace{0.1mm}
    {} $\cls$ {\em Text-Question} $\sep$  {\em Passage} $\sep$\\
    \midrule
        
	\textbf{Conceptual Audio-Text Input} \\
	\vspace{0.1mm}
    {} $\cls$ {\em Speech-Question} $\sep$ {\em Text-Question option} $\sep$ {\em Passage}  $\sep$\\
	\midrule

    \textbf{Question-Discrimination  Input} \\
	\vspace{0.1mm}
    {} $\cls$ {\em \textit{How does scholars divide the library?} } $\sep$ {\em What is the library for ?}
    $\sep$ {\em The Vatican \textbf{at the stella clyde prairie}, more commonly called the Vatican Library or simply the \textbf{fact}, is the library of the Holy See, located in Vatican City...} $\sep$\\
	\midrule
	\textbf{Span Deletion Input} \\
	\vspace{0.1mm}
	{} $\cls$ {\em \textit{How does scholars divide the library?} } $\sep$ {\em What is} $\del$ {\em  for ?}
    $\sep$ {\em The Vatican \textbf{at the stella clyde prairie}}, $\del$ {\em commonly called the Vatican Library or simply the \textbf{fact}}, $\del$ {\em  library of the Holy See, located in Vatican City...} $\sep$\\
    \midrule
    \textbf{Span Substitution  Input} \\
	\vspace{0.1mm}
	{} $\cls$ {\em \textit{How does scholars divide the library?} } $\sep$ {\em What is the library  for ?}
    $\sep$ {\em The Vatican \textbf{at the stella clyde prairie}},  {\em more commonly \textcolor{blue}{named} the Vatican Library or simply the \textbf{fact}}, {\em is the library of the Holy See, \textcolor{blue}{lied} in Vatican City...} $\sep$\\
	\bottomrule	
\end{tabular}
}
\caption{Examples of audio-text input of our model. Original text input is used in a traditional BERT-liked model, \textit{question discrimination} input is used in our self-supervised learning stage, and \textit{span deletion} and \textit{span substitution} inputs are used in a contrastive learning stage. Note that, for the readability, we do not use sub-word tokens in these examples. \textbf{Bold} denotes words in which the ASR error occurs. \textcolor{blue}{Blue} and $\del$ represent the words in which the contrastive learning strategy is used.}
\label{tab:masked_input}
\end{table}

\subsection{Contrastive Learning}
Recent work \cite{wu2020clear} suggests two main arguments: (1) some deletion of unnecessary words in an utterance may not affect the original semantic meaning; (2) suppose that some necessary words (\eg, not) are mistakenly deleted at times, it will provide extremely different semantic meaning. However, injecting some noises (\eg, properly deleting some words) can improve the robustness of the model. Thus, in order to learn effective noise-invariant representation in sentence-level, we train our SQA model with a contrastive objective for performance improvement, in which we augment the training data with two sentence-level augmentation strategies, \textit{span deletion} and \textit{span substitution}\footnote{The two augmentation strategies can happen in any position of the input.}. The augmented input examples are shown in Figure \ref{fig:contras}. More training examples of contrastive learning can be found in Table \ref{tab:masked_input}.

\begin{itemize}
\setlength\itemsep{0.01em}
\item \textbf{Span Deletion}: we add one special token $\del$ to replace the deleted consecutive words of the utterance (\eg, we randomly delete 5 spans, where each is of 5$\%$ length of the textual input sequences).
\item \textbf{Span Substitution}: We randomly sample some words, and then replace them with synonyms to produce the augmented version (\eg, we randomly select 30$\%$ spans of the utterances, and replace them with tokens which share similar semantic meanings).
\end{itemize}

In this stage, we first extract the $\cls$ token representation $H\in R^ {k\times{d}}$ from the last layer of the PLM, where $d=768$ is the dimension of each word vector\footnote{Similar to \cite{kuo2020audio}, we use the $\cls$ token to represent the sentence representation.}. We create augmentations of original utterances with two sentence-level auxiliary tasks on top of the \textit{Question Discrimination}, and then encode the augmented data using the same PLM, used in SQA section (See Figure \ref{fig:my_label}~(a)), to construct the encoded representation $H_{anchor}\in R^ {1\times{d}}$. Our contrastive learning scheme consists of the following components: (1) 
we consider the representation corresponding to the correct $Q_t$ as a \textit{positive}, and others as many \textit{negative}; (2) we use dot-production operation to compute the similarity scores between the joint speech-text representations and the anchor representation; (3) we apply a softmax function to the measured similarity scores. We leverage speech and text data for contrastive training, where the contrastive loss is as follows:
\begin{equation}
    \mathcal{L}_{con}\!=\!-\sum_{i}^{k}y_i\log(\mathrm{softmax}(H\times H_{anchor}^{T}))	
    \label{eq:loss_contrast}
\end{equation}

\paragraph{Multi-Task Learning Setup.}
We optimize our model with two main stages: (1) self-supervised training; (2) contrasitive learning. In the self-supervised training stage, we train our SQA model with three auxiliary tasks to obtain a better local optimum. We use binary cross-entropy loss in all proposed auxiliary tasks. The loss is computed by summing SQA answer prediction loss and all three auxiliary SSL task losses with same ratio. In contrastive learning trainig stage, the loss is defined as a linear combination of SQA answer prediction loss and contrastive loss with the same ratio.

\begin{table*}[t]
\centering

\vspace{-5pt}
\resizebox{0.98\textwidth}{!}{%
\begin{tabular}{l c c c c c c c c}
\toprule
\multirow{2}{*}{\textbf{Method}} &\multirow{2}{*}{\textbf{Spoken-SQuAD}} &\multicolumn{6}{c}{\multirow{2}{*}{\textbf{Spoken-CoQA}}} & \multirow{2}{*}{\textbf{FGC}}
\\ \\ \cmidrule{2-9}
& \bf Overall &\bf Child. &\bf Liter. & \bf Mid-High. &\bf News &\bf Wiki & \bf Overall & \bf Acc \\
\midrule
FlowQA \cite{huang2018flowqa} & 56.7/70.8&22.6/35.8&22.4/35.2&22.0/34.2&21.4/33.6&22.0/34.7&22.1/34.7 & -\\
BERT \cite{devlin2018bert} &58.6/71.1&41.7/55.6&40.1/54.6&39.8/52.7&40.1/53.8& 40.6/53.8&40.6/54.1&77.0\\
BERT + SLU \cite{serdyuk2018towards}  &59.3/71.7& 42.0/55.7&41.4/54.6&40.0/53.1& 40.5/54.0& 41.1/54.6&41.0/54.4&77.6\\
\citet{su2020improving}&59.8/72.6 &42.1/56.0&42.0/56.3&40.0/53.1&40.4/54.0&40.2/54.0&40.9/54.7&77.8\\
BERT+ TS-Attention \cite{kuo2020audio}&59.7/72.4 &42.6/56.6&42.7/56.7&40.3/53.9&41.0/55.0&40.6/54.8&41.6/55.4&78.2\\
\midrule
\multicolumn{8}{l}{\textit{Only using Self-supervised Learning}}\\
BERT + \textsc{UR}
&59.4/71.7&42.6/55.8&41.9/55.6&40.6/53.8&40.9/54.0&40.7/54.3&41.5/54.7 &77.5\\
BERT + \textsc{UI} &59.5/71.9&42.7/55.8&42.3/55.7&41.1/53.9&41.0/54.2&41.2/54.6&41.5/54.8&77.6\\
BERT + \textsc{QD} &59.9/72.4&43.0/56.2&42.2/55.7&41.2/54.3&41.5/54.4&41.6/54.8&41.9/55.1&78.0\\
BERT + \textsc{UR} + \textsc{UI} &59.8/72.6&43.1/56.3&42.3/55.7&41.5/54.5&41.4/54.6&41.5/54.9&41.9/55.2&78.1\\
BERT + \textsc{UR} + \textsc{QD} &60.2/72.6&43.4/56.7&42.6/55.9&41.8/54.7&41.5/54.9&42.0/55.4&42.5/55.5&78.4\\
BERT + \textsc{UI} + \textsc{QD} &60.5/73.0 &43.5/56.8&42.5/56.1&41.6/55.0&41.2/54.8&42.0/55.6&42.4/55.6&78.5\\
BERT + \textsc{UR} + \textsc{UI} + \textsc{QD} &61.0/73.6 &43.9/57.4&42.8/56.7&42.1/55.3&41.9/55.3&42.0/56.0&42.7/56.1&78.8\\
\midrule
\multicolumn{8}{l}{\textit{Only using Contrastive Learning}}\\
BERT + \textsc{SD} &59.2/71.5&42.8/55.5&42.0/55.3&40.5/53.4&40.8/54.0&41.2/54.3&41.5/54.5&77.3\\
BERT + \textsc{SS} &59.4/71.5&42.9/55.7&42.1/55.6&40.3/53.4&41.0/54.1&41.4/54.2&41.5/54.6&77.4\\
BERT + \textsc{SD} + \textsc{SS} &59.6/71.8&43.3/56.1&42.4/55.6&41.2/54.2&41.4/54.5&41.2/54.5&41.9/54.9&77.9\\
\midrule
BERT + T-A Attention &60.3/73.2 &43.0/57.3&42.5/56.1&40.9/55.0&41.9/55.1&41.7/55.5&42.0/55.8&78.7\\
\midrule
\textbf{Ours}&\textbf{62.5/75.5} &\textbf{46.5/59.5}&\textbf{46.1/59.1}&\textbf{44.3/57.3}&\textbf{44.9/57.6}&\textbf{45.2/58.0}&\textbf{45.4/58.3}&\textbf{81.3}\\
\bottomrule
\end{tabular}%
}
\vspace{-5pt}
\caption{The comparison between our method and other method on the SQA performance. \textsc{UR}, \textsc{UI}, and \textsc{QD} denote \textit{utterance resorting}, \textit{utterance insertion}, and \textit{question discrimination}, respectively. \textsc{SD} and \textsc{SS} are \textit{span deletion} and \textit{span substitution}. T-A Attention denotes Temporal-Align Attention. }
\vspace{-10pt}
\label{table:domain}
\end{table*}

\section{Experiments}
In this section, we conduct experiments to compare our proposed method with various baselines and state-of-the-art approaches.

\subsection{Datasets}
We evaluate our approach on three benchmark datasets: Spoken-SQuAD \cite{li2018spoken}, Spoken-CoQA \cite{you2020data}, and FGC\footnote{https://fgc.stpi.narl.org.tw/activity/techai2018}. 

\paragraph{Spoken-SQuAD.}
Spoken-SQuAD \cite{li2018spoken}\footnote{ In original Spoken-SQuAD dataset, questions are in text form. In this work, we utilize Google TTS to translate them into spoken form. } is a large listening comprehension corpus, where the training set and testing set consist of 37k and 5.4k question-answer pairs, respectively. The word error rate (WER) is around 22.77$\%$ in the training set, and around 22.73$\%$ on the testing set. The documents are in the form of speech, and the questions and answers are in the form of text, respectively. The manual transcripts of Spoken-SQuAD are collected from SQuAD benchmark dataset \cite{rajpurkar2016squad}.

\paragraph{Spoken-CoQA.}
Spoken-CoQA~\cite{you2020data} is a large spoken conversational question answering (SCQA) corpus, where the training set and testing set consist of 40k and 3.8k question-answer pairs from 7 multiple domains, respectively. The WER is around 18.7$\%$. The questions and passages are both in the form of text and speech, and answers are in the form of text, respectively. The goal is to generate a time span in the spoken multi-turn dialogues, and then answer questions based on the given passage and conversations.

\paragraph{FGC.}
FGC is a Chinese spoken multi-choice question answering (MCQA) corpus across a variety of domains. The number of question-answer pairs in the training set and testing set is 40k and 3.8k, respectively. Each PQC pair is composed of 1 passage, 1 question, and 4 corresponding answers, where only one answer is correct. All passages, questions, and multiple choices are in spoken form. Following the widely used setting in \cite{kuo2020audio}, we apply the Kaldi toolkit to construct the ASR module. The WER is around 20.4$\%$.


\subsection{Implementation and Evaluation Setup}
We utilize Pytorch to implement our model. We adopt BERT-base as our backbone encoder, which consists of 12 transformer layers. We set the maximum sequence length of input and the hidden vector dimension to 512 and 768, respectively. $k$ in Section 3 is set to 9. We train our model on 2x 2080Ti for 2-3 days with a batch size of 4 per GPU using the Adam optimizer with an initial learning rate of $3\times 10^{-5}$. For Spoken-CoQA, in order to utilize conversation history, we add the current question with previous 2 rounds of questions and ground-truth answers. When trained on FGC, we follow the standard multi-choice setting \cite{kuo2020audio}, which takes questions, each candidate answers, and passages as inputs. We evaluate our model using the  Exact Match (EM) and  F1 to measure the performance of SQA models on Spoken-CoQA and Spoken-SQuAD, following previous work \cite{li2018spoken,kuo2020audio,su2020improving}. For FGC, we choose accuracy to evaluate the model performance on response quality.

\begin{figure*}[t]
    \centering
    \includegraphics[width=0.95\linewidth]{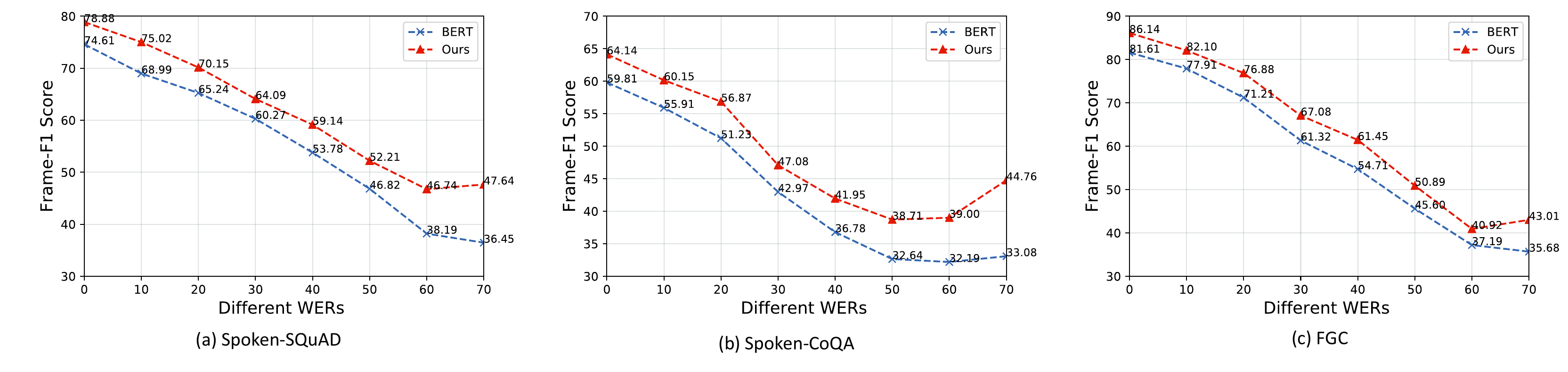}
    \vspace{-10pt}
    \caption{Performances of different WERs.}
    \label{fig:wer}
    \vspace{-10pt}
\end{figure*}

\subsection{Results}

We report quantitative results on Spoken-SQuAD, Spoken-CoQA, and FGC datasets in Table \ref{table:domain}. In our experiments, we set three aspects to study the effectiveness of key components of our method: (1) only using self-supervised learning strategies; (2) only using contrastive learning strategies; (3) we train the model with Temporal-Alignment Attention. Based on these initial aspects, we explore how effective each key component is for SQA.

We first evaluate if the model with three auxiliary tasks can generate a proper answer and how much improvement it can achieve over all evaluated models. For all datasets, our model significantly outperforms all evaluated methods on most of the metrics. Specifically, we observe that sequentially incorporating three proposed strategies brings superior performance improvements in terms of F1 and EM scores. Table \ref{table:domain} compares the importance of different auxiliary SSL tasks, which shows that \textsc{QD} $>$ \textsc{UI} $>$ \textsc{UR} in terms of response quality. This suggests that the auxiliary tasks can effectively aid the learning of the SQA model to learn more sequential information and cross-modality representations for the answer prediction. 

We then compare our method with other methods in terms of contrastive loss on three datasets. In Table \ref{table:domain}, we utilize the proposed contrastive learning with the speech-text input as the auxiliary task, which consistently brings additional performance improvements on all datasets. When further explore the effectiveness of two augmentation strategies, we see that the model achieves comparable performances using \textsc{SD} or \textsc{SS}, and combining both of them enhances the capacity of the model to tackle many unseen sentence pairs. This indicates the importance of noise-invariant representations in boosting performance.

To validate the effectiveness of the proposed T-A Attention, we compare the models with T-A Attention and without it. The model with T-A Attention consistently shows remarkable performance improvements by 60.3$\%$/73.2$\%$ (vs. 58.6$\%$/71.1$\%$) and 42.0$\%$/55.6$\%$ (vs. 40.6$\%$/54.1$\%$) in terms of EM/F1 scores on Spoken-SQuAD and Spoken-CoQA, and 78.7$\%$ (vs. 77.0$\%$) in terms of standard accuracy on FGC. Table \ref{table:domain} shows that our model achieves best results by 62.5$\%$/75.5$\%$ (vs. 58.6$\%$/71.1$\%$), 45.4$\%$/58.3$\%$ (vs. 40.6$\%$/54.1$\%$), and 81.3$\%$ (vs. 77.0$\%$) across three datasets. This suggests that, by taking advantage of the proposed training scheme and  T-A Attention, our model provides a more fine-grained understanding of spoken content to benefit the SQA answer prediction.

\section{Ablation Study}

\paragraph{Effects of Word Error Rates.}
To study how word error rates (WERs) will influence the model performance, we experiment with BERT, which is our baseline model, under different WERs. We randomly split three datasets into small-scale subsets of roughly equal training data size under different WERs for the ablation study. Then we compute Frame-level F1 score~\cite{chuang2019speechbert} to evaluate the robustness of our proposed method with different WERs in Figure~\ref{fig:wer}. We find that our model consistently achieves better results compared to the evaluated baseline. In addition, we find that higher WER leads to a consistent drop in all three spoken question answering tasks. This suggests low WER brings these gains in all SQA settings.

\begin{figure}
    \centering
    \includegraphics[width=0.75\linewidth]{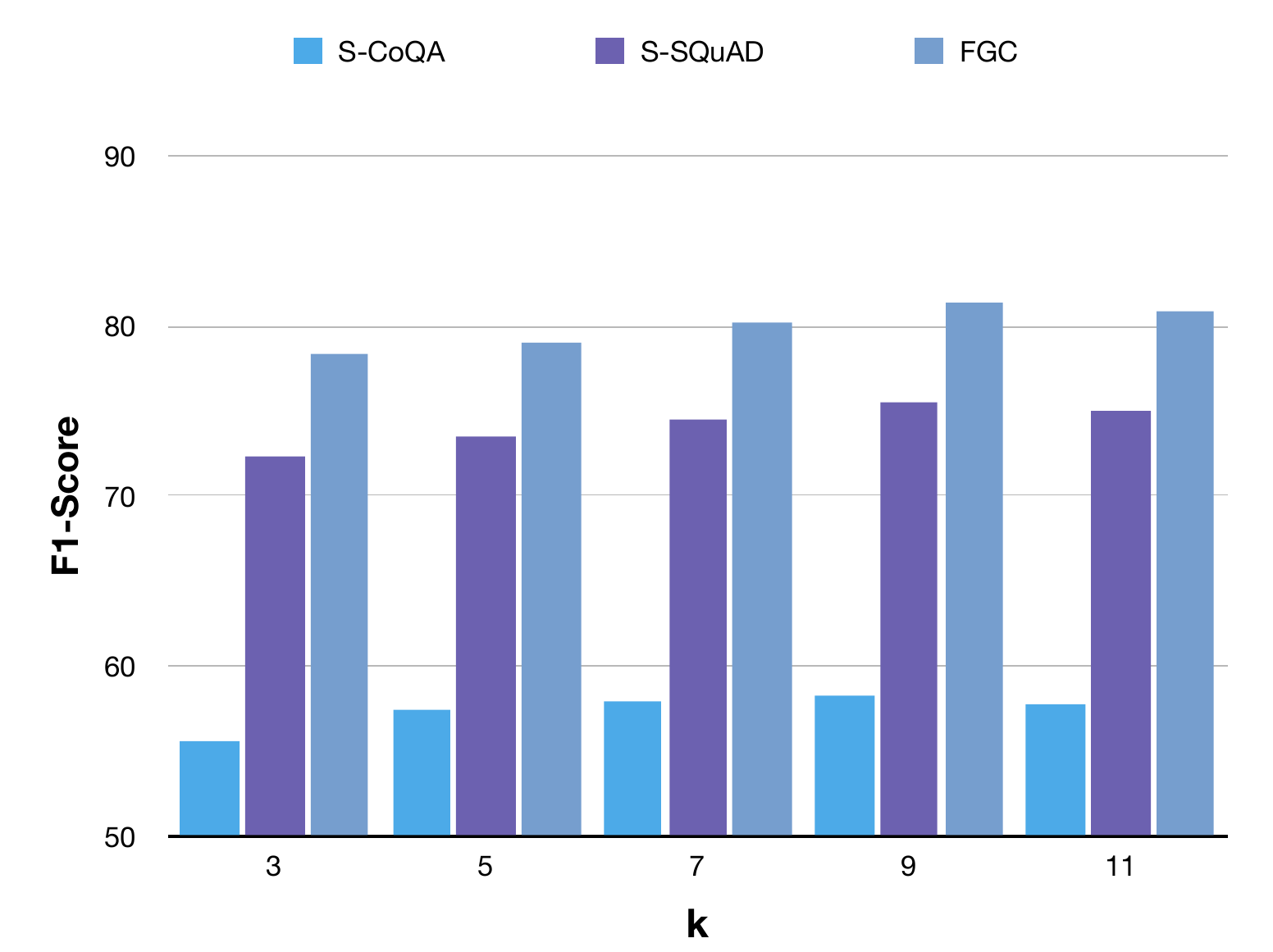}
    \vspace{-10pt}
    \caption{Effects of $k$.}
    \label{fig:Kvale}
    \vspace{-10pt}
\end{figure}

\paragraph{Effects of Hyperparameter Selection.}
Self-supervised training enables the SQA model to capture sequential dependency between utterances along with semantic matching and maintain dialog coherence within a context. We explore the effects of different $k$, which determines the length of utterances in these auxiliary tasks. Figure \ref{fig:Kvale} compares the performance of model  with different $k$. We find that increasing the value of $k$ clearly improves model performance, but it will not further increase after $k$ = 9. We hypothesize that it gives rise to two potential reasons: (1) if the utterance length is too small within the context, the model cannot capture enough contextual information; (2) if the utterance length is too large, which introduces additional noise, it will not benefit the model performance. In our final models, we use $k$ = 9 for self-supervised training.

\begin{table}[t]
\footnotesize
\centering
\resizebox{1.0\columnwidth}{!}{%
\begin{tabular}{lccc}
\toprule
\multirow{2}{*}{\textbf{Algorithm}} &      \textbf{S-SQuAD}     & \textbf{S-CoQA}  &      \textbf{FGC}                \\ \cmidrule{2-4}
             & \multicolumn{1}{c}{F1} & \multicolumn{1}{c}{F1} & \multicolumn{1}{l}{Acc. (\%)} \\ \midrule
          BERT   &71.1&54.1&77.0\\
           
          \quad  w/ Co-Att \cite{lu2019vilbert}             &      72.8               &     55.0                &   77.9                   \\ 
          \quad  w/ ICCN \cite{sun2020learning}             &       71.7                &            54.7         &         77.7            \\ 
          \quad  w/ S-Fusion \cite{siriwardhana2020jointly}            & 68.1                    &     51.8            &  75.1
             \\ 
            \midrule
          \quad w/   ST-Attention         &      \textbf{73.2}                 &        \textbf{55.8}              &        \textbf{78.7}             \\
            \bottomrule
\end{tabular}
}
\vspace{-5pt}
\caption{Effect of different attention mechanism.}
\vspace{-10pt}
\label{tab:attention}
\end{table}

\paragraph{Effects of T-A Attention.}
We further evaluate the effectiveness of various attention mechanisms in Table \ref{tab:attention}. We define BERT as the base model. We observe that the model with the proposed T-A attention strategy achieves state-of-the-art performance on three datasets. It clearly demonstrates T-A attention can effectively reduce the discrepancy between text and speech domains.

\section{Conclusions}
Spoken question answering requires fine-grained understanding of both speech and text data. To this end, we propose a novel training scheme for spoken question answering. By carefully designing several auxiliary tasks, we incorporate the self-supervised contrastive learning framework to capture consistency and coherence within speech documents and text corpus without any additional data. We further propose a novel Temporal-Alignment strategy to align audio features and textual concepts by performing mutual attention over two modalities. Our model achieves state-of-the-art performance on three SQA benchmark datasets. For future work, we will develop more effective auxiliary tasks to enhance the quality of answer prediction.

\bibliography{custom}
\bibliographystyle{acl_natbib}

\appendix
\section*{Appendix}
\section{More Examples}
Table \ref{table:task1ex-canada} show examples used in the self-supervised training stage.

\begin{table*}[!tbp]
\centering
\resizebox{0.9\textwidth}{!}{%
\begin{tabular}{p{4cm}p{13.5cm}}
\toprule
\textbf{ASR Passage Title} & \textbf{Vatican-Library}\\

\midrule
\textbf{ASR Question} & How does scholars divide the library? \\
\midrule

\textbf{Original ASR Content}  &

The Vatican \textbf{at the stella clyde prairie}, more commonly called the Vatican Library or simply the \textbf{fact}, is the library of the Holy See, located in Vatican City. Formally established in 1475, although it is much older, it is one of the oldest libraries in the world and contains one of the most significant collections of historical \textbf{tax}. It has 75,000 \textbf{courtesies} from throughout history, as well as 1.1 million printed books, which include some 8,500 \textbf{king abdullah}. The Vatican Library is a research library for history, \textbf{lot}, philosophy, science and theology. The Vatican Library is open to anyone who can document their qualifications \textbf{in} research needs. Photocopies for private study of pages from books published between 1801 and 1990 can be requested in person or by mail. In March 2014, \textbf{team} the Vatican Library began an initial four-year project of digitising its collection of manuscripts, to be made available online. The Vatican Secret Archives were separated from the library at the beginning of the 17th century; they contain another 150,000 items. Scholars have traditionally divided the history of the library into five periods, \textbf{pre ladder and ladder and having yon prevent a cannon vatican}. The \textbf{pre latter in period}, comprising the initial days of the library, dated from the earliest days of the Church. Only a handful of volumes survive from this period, \textbf{the summer} very significant.
\\
\midrule 
\textbf{Utterance Insertion}

& 
The Vatican \textbf{at the stella clyde prairie}, more commonly called the Vatican Library or simply the \textbf{fact}, is the library of the Holy See, located in Vatican City. Formally established in 1475, although it is much older, it is one of the oldest libraries in the world and contains one of the most significant collections of historical \textbf{tax}. It has 75,000 \textbf{courtesies} from throughout history, as well as 1.1 million printed books, which include some 8,500 \textbf{king abdullah}. The Vatican Library is a research library for history, \textbf{lot}, philosophy, science and theology. The Vatican Library is open to anyone who can document their qualifications \textbf{in} research needs. Photocopies for private study of pages from books published between 1801 and 1990 can be requested in person or by mail. \textcolor{blue}{The highly prized memorabilia which included item spanning the many stages of jackson's courier came for more than thirty fans associates and family members who contacted julian factions to sell their gifts and mementos of the singer.}
In March 2014, \textbf{team} the Vatican Library began an initial four-year project of digitising its collection of manuscripts, to be made available online. The Vatican Secret Archives were separated from the library at the beginning of the 17th century; they contain another 150,000 items. Scholars have traditionally divided the history of the library into five periods, \textbf{pre ladder and ladder and having yon prevent a cannon vatican}. The \textbf{pre latter in period}, comprising the initial days of the library, dated from the earliest days of the Church. Only a handful of volumes survive from this period, \textbf{the summer} very significant.
\\  
\midrule 
\textbf{Utterance Restoration}  & 
The Vatican \textbf{at the stella clyde prairie}, more commonly called the Vatican Library or simply the \textbf{fact}, is the library of the Holy See, located in Vatican City. Formally established in 1475, although it is much older, it is one of the oldest libraries in the world and contains one of the most significant collections of historical \textbf{tax}. It has 75,000 \textbf{courtesies} from throughout history, as well as 1.1 million printed books, which include some 8,500 \textbf{king abdullah}. \textcolor{blue}{\mask, \mask, \mask, \ldots, \mask.} Photocopies for private study of pages from books published between 1801 and 1990 can be requested in person or by mail. In March 2014, \textbf{team} the Vatican Library began an initial four-year project of digitising its collection of manuscripts, to be made available online. The Vatican Secret Archives were separated from the library at the beginning of the 17th century; they contain another 150,000 items. Scholars have traditionally divided the history of the library into five periods, \textbf{pre ladder and ladder and having yon prevent a cannon vatican}. The \textbf{pre latter in period}, comprising the initial days of the library, dated from the earliest days of the Church. Only a handful of volumes survive from this period, \textbf{the summer} very significant.
\\
\bottomrule 
\end{tabular}
}
\caption{Example of \textit{Utterance Insertion} and \textit{Utterance Restoration}. \textbf{Bold} denotes the words in which the ASR error occurs. \textcolor{blue}{Blue} and $\mask$ are the words in which the self-supervised learning strategies are used.}
\label{table:task1ex-canada}
\end{table*}

\end{document}